\documentclass[lefttitle,twoside,final,10pt,5p,twocolumn,authoryear]{elsarticle_modified}

\usepackage[osf,sc]{mathpazo}
\usepackage[scaled=0.85]{FiraMono} 
\usepackage[scaled=.95]{cabin} 
\usepackage{newpxmath} 

\RequirePackage{microtype}

\usepackage{sectsty}

\usepackage{ragged2e}

\usepackage{amssymb}

\usepackage{amsmath}

\usepackage{longtable}
\numberwithin{table}{section}

\usepackage{xcolor}
\definecolor{cobalt}{rgb}{0.0, 0.28, 0.67}
\usepackage[colorlinks]{hyperref}
\AtBeginDocument{\hypersetup{
    colorlinks=true,
    linkcolor=cobalt,
    filecolor=black,
    citecolor=cobalt,
    urlcolor=cobalt,
    }}

\newcommand*{\doi}{}
\makeatletter
\newcommand{\doi@}[1]{\ttfamily\href{https://doi.org/#1}{#1}}
\DeclareRobustCommand{\doi}{\hyper@normalise\doi@}
\makeatother

\usepackage{booktabs}

\usepackage{dirtytalk}

\makeatletter
\def\runauthor{G. Puebla and L. A. A. Doumas}
\def\runtitle{Preprint: Learning Rules from Rewards}
\makeatother

\begin{document}
\allsectionsfont{\sffamily}

\begin{frontmatter}

\author[label1]{Guillermo Puebla\corref{cor1}}
\cortext[cor1]{Corresponding author. Email: \texttt{pueblaramirezg@gmail.com}}
\affiliation[label1]{
    organization={Facultad de Administración y Economía, Universidad de Tarapacá},
    country={Chile}
             }
\author[label2]{Leonidas A. A. Doumas}
\affiliation[label2]{organization={School of PPLS, University of Edinburgh},
             country={United Kingdom}
             }

\title{\textsf{\huge Learning Rules from Rewards}} 

\begin{abstract}
\justifying
Humans can flexibly generalize knowledge across domains by leveraging structured relational representations. While prior research has shown how such representations support analogical reasoning, less is known about how they are recruited to guide adaptive behavior. We address this gap by introducing the Relational Regression Tree Learner (RRTL), a model that incrementally builds policies over structured relational inputs by selecting task-relevant relations during the learning process. RRTL is grounded in the framework of relational reinforcement learning but diverges from traditional approaches by focusing on ground (i.e., non-variabilized) rules that refer to specific object configurations. Across three Atari games of increasing relational complexity (Breakout, Pong, Demon Attack), the model learns to act effectively by identifying a small set of relevant relations from a broad pool of candidate relations. A comparative version of the model, which partitions the state space using relative magnitude values (e.g., \say{more}, \say{same}, \say{less}), showed more robust learning than a version using logical (binary) splits. These results provide a proof of principle that reinforcement signals can guide the selection of structured representations, offering a computational framework for understanding how relational knowledge is learned and deployed in adaptive behavior.
\end{abstract}

\end{frontmatter}



\section{Introduction}\label{intro}

Humans possess the remarkable ability to generalize knowledge flexibly and rapidly across domains, a capacity that surpasses even the most powerful artificial intelligence (AI) systems \citep{hummel2025basic,greff2020binding,mitchell2021abstraction,chollet2025arc}. Previous research has shown that this ability, known as cross-domain generalization, helps us when faced with formal problems, such as solving physics problems \citep{bassok1989interdomain,cooper1987effects} or understanding scientific concepts \citep{gick1983schema,donnelly1993use,gentner1983flowing,gentner1983structure}. Beyond these formal settings, this ability also helps us when learning skills to interact effectively with our environment. For example, someone who knows how to ride a bike might find it easier to learn to ride a scooter due to the similar balance and coordination challenges involved. \citet{doumas2022theory} proposed that cross-domain generalization is best understood as a form of analogical inference over structured relational representations. This proposal was instantiated as a computational model based on the DORA framework for relational representation learning \citep{doumas2008theory}. This work demonstrated that a computational system can learn relational invariants from raw visual input, learn structured (i.e., symbolic) representations of those invariants, and then generalize across tasks\textemdash such as from one video game to another\textemdash via role-based relational reasoning (i.e., analogy).

However, while this work offered a compelling account of how relational representations are learned and used to make analogical inferences, it left an important question open: How does a system equipped with a rich vocabulary of structured relations learn to act in the world based on those representations? That is, once we have learned a vocabulary of relational concepts, how do we learn which of these to apply in a given situation? This issue is critical since human beings acquire a vast vocabulary of relations throughout their lives that can apply freely to any given situation \citep{gentner2017analogy}, and each real-world situation consists of a multitude of possible relations to represent, resulting in a combinatorial explosion of representations even for scenarios of moderate complexity \citep{halford1998processing}. Therefore, a central problem for any theory that posits a relational representation of the environment is to determine which of all available relations are relevant for a given task and how to use them to guide adaptive behavior.

In the present work, we take a step toward bridging this gap by developing a simple model of relational policy learning that draws on insights from relational reinforcement learning (RRL), a subfield of symbolic machine learning concerned with learning policies over relational structures \cite[for a review see][]{otterlo2012solving}. In brief, we show how a system can learn relational rules incrementally through interaction with the environment. Our model utilizes a regression tree-based function approximator to learn relational policies in a bottom-up manner, without requiring the prior specification of the most relevant relations for each task. We evaluate our model in a series of simulations using three Atari games (Breakout, Pong, and Demon Attack) that vary in relational complexity. Across these environments, the model learns to build effective relational policies from a broad candidate set of relations, demonstrating that relational structure can serve not only as a medium for generalization but also, crucially, as a substrate for action selection through reinforcement. 

In the following, we provide a brief overview of reinforcement learning (RL) and RRL. We then describe our model, a relational regression tree learner capable of selecting relevant relations and building policies incrementally. Next, we present a series of simulations across three Atari games (Breakout, Pong, and Demon Attack) that vary in relational complexity. Finally, we discuss the implications of our findings, the relationship of our model to existing theories of relational reasoning and learning, and directions for future research. 

\subsection{Reinforcement Learning}

In RL an agent interacts with an environment by taking actions to maximize cumulative rewards. The environment is formalized as a set of states with transitions between them probabilistically determined by the agent's actions. RL algorithms aim to learn an optimal \emph{policy} (a mapping between states and actions) through this interaction \citep{sutton2018reinforcement}. In general terms, RL algorithms can be classified into \emph{model-based} and \emph{model-free} methods. Model-based methods involve learning a transition model and a reward function, which are used to plan the best course of action at a given state through simulation. In contrast, model-free methods use prediction errors to directly learn the \emph{value} (expected cumulative rewards) of taking each action in each state, without learning a model of the environment. These values can then be used to build the policy by greedy selection of the best action in each state. The present work focuses on model-free learning. We use the classic \emph{Q-learning} algorithm \citep{watkins1989learning}, which estimates the expected cumulative future rewards of taking an action in a given state. Q-values are updated using the rule:

\begin{align}\label{eq:1}
    Q(S_{t},A_{t})\leftarrow~& Q(S_{t},A_{t}) \nonumber\\ 
    &  + \alpha \Big[R_{t} + \gamma \max _{a}Q(S_{t+1},a) - Q(S_{t},A_{t})\Big]
\end{align}

where $Q(S_{t},A_{t})$ is the current value of taking action $A$ on state $S$, $\alpha$ is the learning rate, $\gamma$ is a discount factor that has the effect of weighting more rewards closer in time to rewards farther away in the future, and the subtraction term is the prediction error. Although Q-learning evaluates the best next action, the agent may follow a different (e.g., exploratory) policy during training. In this work, we use an $\epsilon$-greedy exploration strategy: the agent selects the action with the highest Q-value most of the time but chooses a random action with probability $\epsilon$, which decays over time to encourage convergence to optimal behavior.

While Q-learning is guaranteed to converge to the optimal policy as long as all state-action pairs are updated during learning \citep{watkins1992q}, it becomes prohibitively expensive for large state spaces. This is especially problematic in relational settings, where the size of the state space grows combinatorially with the number of relations and objects \citep{van2009logic}. As explained in the next section, RRL algorithms use specialized function approximators to handle this problem. 

\subsection{Relational Reinforcement Learning}

The goal of RRL is to learn an optimal policy in an environment described as a set of objects and their relations \citep{dvzeroski2001relational}. Notably, the policy is usually represented as a set of \emph{variabilized} rules in (a subset of) first-order logic \citep[e.g.,][]{driessens2001speeding,guestrin2003generalizing,pasula2007learning}. For example, in a version of the classic problem known as \emph{blocks world} \citep{slaney2001blocks}, where the agent is rewarded for unstacking a group of blocks, one such rule could be: $\texttt{move}(X, \texttt{floor}) \leftarrow \texttt{on}(X,Y) \wedge \texttt{top}(X)$ (i.e., \say{if block $X$ is on any other block $Y$, and block $X$ is on top of a pile, then move block $X$ to the floor}). In contrast, the present work concentrates on learning \emph{ground} rules, i.e., rules that apply to specific objects in a specific task, such as: $\texttt{LEFT} \leftarrow \texttt{more}\textnormal{-}\texttt{x}(\texttt{player},\texttt{ball})$ (i.e., \say{if the player is to the right of the ball, then move left}). We think that it is likely that people learn relational rules that apply to specific situations and, later on, those rules are generalized through the process of schema induction \cite[e.g.,][]{gick1983schema, hummel2003symbolic}. 

\begin{figure*}[t!]
    \centering
    \includegraphics[width=\linewidth]{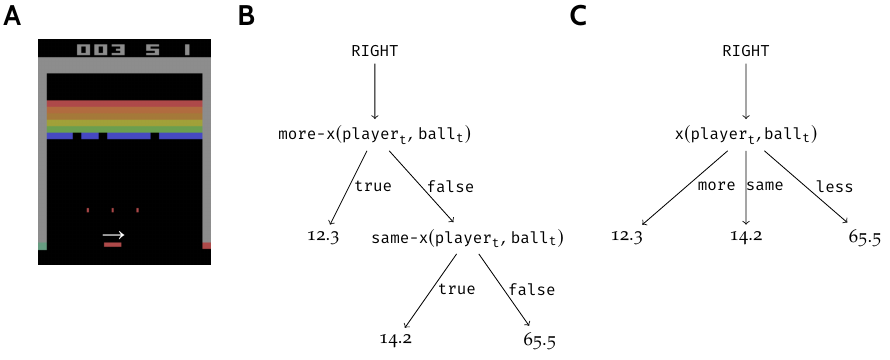}
    \caption{Two different ways of representing state-action values. Panel A shows three potential states of the Breakout environment where the player is either to the left of the ball, at the same x-coordinate, or to the right of the ball. The fact that the value of the action \texttt{RIGHT} is higher when the player is to the left of the ball, lower when the player and the ball are at the same x-coordinate, and lowest when the player is to the right of the ball, can be represented as a logical relational regression tree (Panel B) or as a comparative relational regression tree (Panel C). See text for details.}
    \label{fig:1}
\end{figure*}

From a cognitive point of view, an interesting attribute of classic RRL algorithms is that they build policies \emph{incrementally}, gradually adding rules to the policy that improve its overall quality \citep{driessens2001speeding}. This is in stark contrast to Bayesian theory-based RL \citep{tsividis2021human,pouncy2022inductive}, where the space of possible programs is defined \emph{a priori} and learning equates to inferring the best-fitting program from all possible programs through Bayesian inference. Another important attribute of RRL algorithms is that they have to deal with the discrete nature of relational representations. This is because describing the state of the environment in terms of relations imposes sharp partitions of the state space. For instance, on the aforementioned \texttt{LEFT} rule, the relation $\texttt{more}\textnormal{-}\texttt{x}(\texttt{player},\texttt{ball})$ partitions the state space into states where the player is to the right of the ball, and states where it is not. This fact requires the use of a specialized function approximator that can make use of relational representations to \emph{abstract away} irrelevant aspects of the state space \citep{van2009logic}.

\subsection{Relational Regression Tree Learner}

Our model, which we call the Relational Regression Tree Learner (RRTL), is based on the function approximator for RRL proposed by Driessens \citep{driessens2001speeding,Driessens2004PhD}. This algorithm uses regression trees to represent the state-action value function. For each action, there is a tree where each node represents a conjunction of ground relations, and each leaf is a Q-value. In Driessens's original model, these trees were based on logical splits of the state-action space. To illustrate this point, \autoref{fig:1}a shows three potential states of the game Breakout, where the player gains points by making the ball bounce against the wall of bricks at the top of the screen. In general, the action \texttt{RIGHT} has the highest value when the player is to the left of the ball, has a lower value when the player and the ball are at the same position on the x-axis, and has the lowest value when the player is to the right of the ball. To represent this ranking of values, a logical regression tree needs to make two splits (\texttt{true} and \texttt{false} for \texttt{more}-\texttt{x} and \texttt{true} and \texttt{false} for \texttt{same}-\texttt{x}), as shown in \autoref{fig:1}b. An alternative way of representing the same ranking is to make splits based on the \emph{comparative} values \texttt{more}, \texttt{same}, and \texttt{less} of the \texttt{x} relation between the player and the ball. In this case, the tree needs to make only a single split, as depicted in \autoref{fig:1}c. 

While in principle both kinds of splits can express the same policies, in this work, we propose that comparative splits should lead to more robust learning. There are two reasons for this claim. First, because comparative relations are based on relative magnitude information about a specific dimension\textemdash e.g., a relation like $\texttt{taller}(X,Y)$ is based on an underlying comparison on the dimension \say{height} \citep[see][]{doumas2021model}\textemdash the Q-values of the partitions of the state-action space based on the magnitude categories \say{more}, \say{less} and \say{same} are probably related for most environments. Second, as expressing the same policy using a logical regression tree necessarily involves making more splits in comparison to a comparative regression tree, and every time a split is made there is a chance of choosing an incorrect relation to base the split on, there are more chances of error in the logical regression tree case.

In both cases, to make a prediction, the agent traverses the tree corresponding to each action according to the relations present in the state until it reaches a leaf. The predicted Q-value can then be used to select an action and can be updated according to \autoref{eq:1}. 

At the beginning of the learning process all the trees have a single leaf. At this stage, the Q-value represents the overall value of the action in the environment. All state-action trees consider the same initial set of candidate relations to grow new leaves. As the agent interacts with the environment, each state-action tree keeps track of the current number of visits to the candidate relation, $n$, the mean, $\mu_n$, and the scaled variance, $J_n = \sigma^2_n \cdot n$, of the Q-values produced at each time step, as well as the same statistics for all potential partitions induced by the candidate (i.e., \texttt{true} and \texttt{false} for logical partitions and \texttt{more}, \texttt{same} and \texttt{less} for comparative ones). These statistics are calculated incrementally\footnote{The implementation described by \cite{Driessens2004PhD} uses the sum of squared Q-values to calculate the variance; however, this can be numerically unstable.} according to \autoref{eq:2} and \autoref{eq:3}:

\begin{equation}\label{eq:2}
\mu_{n} = \mu_{n-1} + \frac{x_{n} - \mu_{n-1}}{n}
\end{equation}
\begin{equation}\label{eq:3}
J_{n} = J_{n-1} + (x_{n}-\mu_{n-1})(x_{n}-\mu_n)
\end{equation}

where $\sigma^2_{n} = J_n/n$.

After a \emph{minimal sample size} (a free parameter of the model) has been reached, these statistics can be used to compute, for each candidate, the F-ratio between the variance of the Q-values if the leaf was split according to the candidate and the variance of the Q-values of the unsplit leaf. \autoref{eq:4} and \autoref{eq:5} show the F-ratio for logical and comparative splits, respectively: 

\begin{equation}\label{eq:4}
F = \frac{\frac{n_T}{n_O}\sigma^{2}_T + \frac{n_F}{n_O}\sigma^{2}_F}{\sigma^{2}_O} = \frac{
J_T/n_O+J_F/n_O}{J_O/n_O}
\end{equation}

\begin{equation}\label{eq:5}
F = \frac{\frac{n_M}{n_O}\sigma^{2}_M + \frac{n_S}{n_O}\sigma^{2}_S + \frac{n_L}{n_O}\sigma^{2}_L}{\sigma^{2}_O}=\frac{J_M/n_O+J_S/n_O+J_L/n_O}{J_O/n_O}
\end{equation}

were $\sigma^{2}$ is the variance, $n$ is the total number of visits to the partition, and the subscript now indicates the partition ($T=\textnormal{true}$, $F=\textnormal{false}$, $M=\textnormal{more}$, $S=\textnormal{same}$,  $L=\textnormal{less}$ and $O=\textnormal{overall}$). With this ratio, the tree calculates the $p$-values of a standard one-tailed F-test for all candidates. If the smallest $p$-value is smaller than the \emph{significance level}, the leaf is split according to the candidate, and the process continues until the tree cannot find new splits or reaches a \emph{maximum tree depth}.

\section{Simulations}

\subsection{Environments}

In our simulations we used the environments \say{ALE/Breakout-v5}, \say{ALE/Pong-v5}, and \say{ALE/DemonAttack-v5} of the Gymnasium library \citep{towers2024gymnasium}. As explained below, our chosen games allowed us to test RRTL's capability to handle environments of increasing relational complexity. Our code, which allows exact replication of the results presented here, is available at \url{https://github.com/GuillermoPuebla/rrl}.

\begin{table*}
    \caption{Breakout State Representation}
    \label{tab:BreakoutRelations}
    \centering
    \begin{tabular}{@{}cllll@{}}         
    \toprule
    & & & \multicolumn{2}{c}{Relation} \\
    \cmidrule(l){4-5}
    Dimension & Object-1 & Object-2 & Logical & Comparative \\ 
    \midrule
    
    \texttt{x} & \texttt{player}$_{\texttt{t}}$ & \texttt{ball}$_{\texttt{t}}$ & \texttt{more-x}(\texttt{player}$_{\texttt{t}}$, \texttt{ball}$_{\texttt{t}}$) & \texttt{x}(\texttt{player}$_{\texttt{t}}$, \texttt{ball}$_{\texttt{t}}$)\\
    
    & & & \texttt{same-x}(\texttt{player}$_{\texttt{t}}$, \texttt{ball}$_{\texttt{t}}$) & \\
    
    & & & \texttt{less-x}(\texttt{player}$_{\texttt{t}}$, \texttt{ball}$_{\texttt{t}}$) & \\ \midrule
    
    \texttt{y} & \texttt{player}$_{\texttt{t}}$ & \texttt{ball}$_{\texttt{t}}$ & \texttt{more-y}(\texttt{player}$_{\texttt{t}}$, \texttt{ball}$_{\texttt{t}}$) & \texttt{y}(\texttt{player}$_{\texttt{t}}$, \texttt{ball}$_{\texttt{t}}$)\\
    
    & & & \texttt{same-y}(\texttt{player}$_{\texttt{t}}$, \texttt{ball}$_{\texttt{t}}$) & \\
    
    & & & \texttt{less-y}(\texttt{player}$_{\texttt{t}}$, \texttt{ball}$_{\texttt{t}}$) & \\ \midrule
    
    \texttt{x} & \texttt{ball}$_{\texttt{t}}$ & \texttt{ball}$_{\texttt{t-1}}$ & \texttt{more-x}(\texttt{ball}$_{\texttt{t}}$, \texttt{ball}$_{\texttt{t-1}}$) & \texttt{x}(\texttt{ball}$_{\texttt{t}}$, \texttt{ball}$_{\texttt{t-1}}$) \\
    
    & & & \texttt{same-x}(\texttt{ball}$_{\texttt{t}}$, \texttt{ball}$_{\texttt{t-1}}$) & \\
    
    & & & \texttt{less-x}(\texttt{ball}$_{\texttt{t}}$, \texttt{ball}$_{\texttt{t-1}}$) & \\
    \midrule
    
    \texttt{y} & \texttt{ball}$_{\texttt{t}}$ & \texttt{ball}$_{\texttt{t-1}}$ & \texttt{more-y}(\texttt{ball}$_{\texttt{t}}$, \texttt{ball}$_{\texttt{t-1}}$) & \texttt{y}(\texttt{ball}$_{\texttt{t}}$, \texttt{ball}$_{\texttt{t-1}}$) \\
    
    & & & \texttt{same-y}(\texttt{ball}$_{\texttt{t}}$, \texttt{ball}$_{\texttt{t-1}}$) & \\
    
    & & & \texttt{less-y}(\texttt{ball}$_{\texttt{t}}$, \texttt{ball}$_{\texttt{t-1}}$) & \\ 

    \bottomrule
    \end{tabular}
    
\end{table*}

\subsubsection{Breakout}

This was the simplest environment we used. As previously explained, in Breakout the player controls a paddle and receives points when the ball bounces off the wall at the top of the screen (see \autoref{fig:1}a). The player loses points if the ball passes the paddle when it is going down (the player's y-position is fixed at the bottom of the screen). The actions available to the player are: \texttt{NOOP}, \texttt{FIRE}, \texttt{RIGHT}, \texttt{LEFT}. To succeed in this game, the player needs to learn to follow the ball, which requires paying more attention to the relations across the x-dimension than to the relations across the y-dimension\footnote{There are certainly more complex and effective policies like digging a tunnel through the wall to allow the ball to bounce above the blocks \cite[e.g.,][]{mnih2015human}. However, as we are not representing the wall, following the ball is the optimal policy in the environment represented at this level of abstraction.}. \autoref{tab:BreakoutRelations} presents all relations used to represent the state of the environment\footnote{Importantly, \citet{doumas2022theory} have previously shown that structured (i.e., symbolic) representations of all of the relations used in these simulations are learnable from simple visual inputs.}. As can be seen, we used the \texttt{x} and \texttt{y} relations between the player and the ball and (henceforth, object relations) and the \texttt{x} and \texttt{y} relations between the ball at the current time step and the ball at the previous time step (henceforth, trajectory relations)\footnote{We did not represent the trajectory relations for the player because, as noted above, the y-trajectory of the player is a constant.}. In object relations, the first object was always the player, and the second object was always the ball. In the trajectory relations, the first object was always the object at the current time step, and the second was the object at the previous time step. We created two versions of the model, a \emph{logical} version and a \emph{comparative} version, where we used the relations of the \say{logical} and \say{comparative} columns of \autoref{tab:BreakoutRelations}, respectively. Note that even in this simple environment, tabular Q-learning consistently fails to converge to a policy more effective than random behavior. 

During the construction of the state, we filtered out all states where the ball was not present (i.e., those states were treated as empty). Aditionally, when the state was empty, the agent always took a randomly sampled action. This was done because the frequentist statistics approach used to determine the state-action tree splits requires all candidates to have the same number of visits in order to compete on equal footing.

To obtain a relational state from the Gymnasium environment, we created a visual pre-processor, which used the color and shape of the objects to calculate the x and y positions of the ball and the paddle. To obtain the relations shown on \autoref{tab:BreakoutRelations}, we set the roles \say{Object-1} and \say{Object-2} according to a hierarchy where \say{Object-1} was always the player and \say{Object-2} was always the ball. Furthermore, to calculate the comparative value associated with each relation we subtracted the x and y positions according to the same hierarchy and categorized the difference with a tolerance level of six pixels (i.e., if the player was 6 pixels or more to the right of the ball the comparative value was \texttt{more}, if it was 6 pixels or more the left of the ball the comparative value was \texttt{less} and if the difference was below six pixels the comparative value was \texttt{same}. We calculated the trajectory relations in the same way, except that we used a tolerance level of zero.

\begin{figure}[hbt!]
\centering
\includegraphics[width=0.4\linewidth]{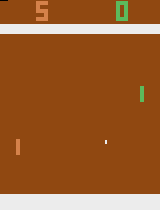}
\caption{A typical state of the Pong environment.}
\label{fig:Pong}
\end{figure}

\subsubsection{Pong}

In this environment, two paddles, one corresponding to the player and the other to the enemy, hit a ball with the objective of hitting it past the opponent (see \autoref{fig:Pong}). The player receives a positive reward when the ball passes the enemy, and a negative reward when the ball passes the player. The episode ends when either the player or the enemy makes the ball pass the other 21 times. Both the player and the enemy can move only on the y-axis, while the ball can move on the x-axis and the y-axis. The actions available to the player in this environment are: \texttt{NOOP}, \texttt{FIRE}, \texttt{RIGHT}, \texttt{LEFT}, \texttt{RIGHTFIRE}, \texttt{LEFTFIRE}. However, as the last two actions had the same effect as \texttt{RIGHT} and \texttt{LEFT}, we omitted them for simplicity. Because in Pong there are three objects instead of two, the number of potential object and trajectory relations increases accordingly. Furthermore, in addition to the object and trajectory relations used in the previous simulation, we introduced two \texttt{contact} relations between the player and the ball, and between the ball and the enemy (these relations are necessarily logical). \autoref{tab:PongTable} in the Appendix presents all the relations used to represent the state of the Pong environment. 

As in Breakout, during the construction of the state, we filtered out all states where the ball was not present. When the state was empty, the agent always took a randomly sampled action.

To obtain a relational state from the Gymnasium environment, we created a visual pre-processor that calculated the x- and y-positions of the objects. We used a tolerance level of 4 pixels to calculate the comparative value associated with each object relation. For the trajectory relations, we used a tolerance level of zero. The following hierarchy of objects was used to represent the state: $\texttt{player} > \texttt{ball} > \texttt{enemy}$ (see \autoref{tab:PongTable}).

\begin{figure*}[ht!]
    \centering
    \includegraphics[width=0.53\linewidth]{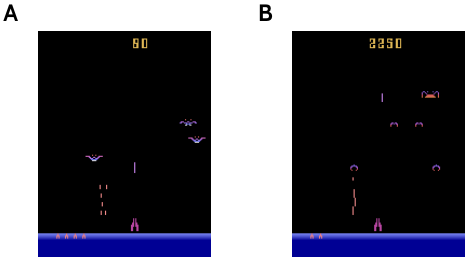}
    \caption{Two typical states of the Demon Attack environment. Panel A shows a state where there are three big enemies that can shoot missiles. Panel B shows a state where, besides big enemies, there are small enemies. The player loses a life if it is touched by a small enemy.}
    \label{fig:DemonAttackState}
\end{figure*}

\subsubsection{Demon Attack}

In this environment, the player controls a spaceship at the bottom of the screen that can only move on the x-dimension (see \autoref{fig:DemonAttackState}). In the initial levels, \emph{big enemies} (or demons) appear in waves of three in the upper part of the screen. The bottom-most enemy shoots projectiles that cause the player to lose a life and receive a negative reward. The player can shoot missiles that destroy the enemies upon contact, producing a positive reward. When an enemy is destroyed, a new one appears to take its place until the current level is completed. Once all the enemies on a particular level are destroyed, the player moves on to the next, more difficult wave. On advanced levels, the big enemies split into two bird-like \emph{small enemies} the first time they are shot. The small enemies will eventually attempt descent onto the spaceship, which will also cause the player to lose a life and receive a negative reward. The actions available to the player in this environment are: \texttt{NOOP}, \texttt{FIRE}, \texttt{RIGHT}, \texttt{LEFT}, \texttt{RIGHTFIRE}, \texttt{LEFTFIRE}. We used all the available actions in this simulation. \autoref{tab:DemonAttackTable} in the Appendix presents all the relations used to represent the state of this environment. Because in Demon Attack there can be up to three big enemies (e.g., \texttt{e-big-1}) and up to six small enemies (e.g., \texttt{e-small-6}) at any given time\footnote{With the constrain that for each pair of small enemies there is one less possible big enemy.}, the number of potential object relations is quite large. In this simulation, we only used object relations between the player and the other objects on the screen, except for the player's missile, which we treated as part of the player's action. Furthermore, we treated the enemy's projectiles as a single object, which we termed \texttt{e}-\texttt{missile}. In this game, we did not use any trajectory relations. This was because the demons follow a non-linear trajectory in the x- and y-dimensions even when stationary, circling a fixed point. 

Similarlly to the previous games, when building the relational state, we filtered out all states where the enemy missile was not present, in which case the agent always took a randomly sampled action.

As with the previous games, we created a visual pre-processor that calculated the x- and y-positions of the objects. We used a tolerance level of 3 pixels to calculate the comparative value associated with each object relation.

\begin{figure*}[ht!]
    \centering
    \includegraphics[width=\linewidth]{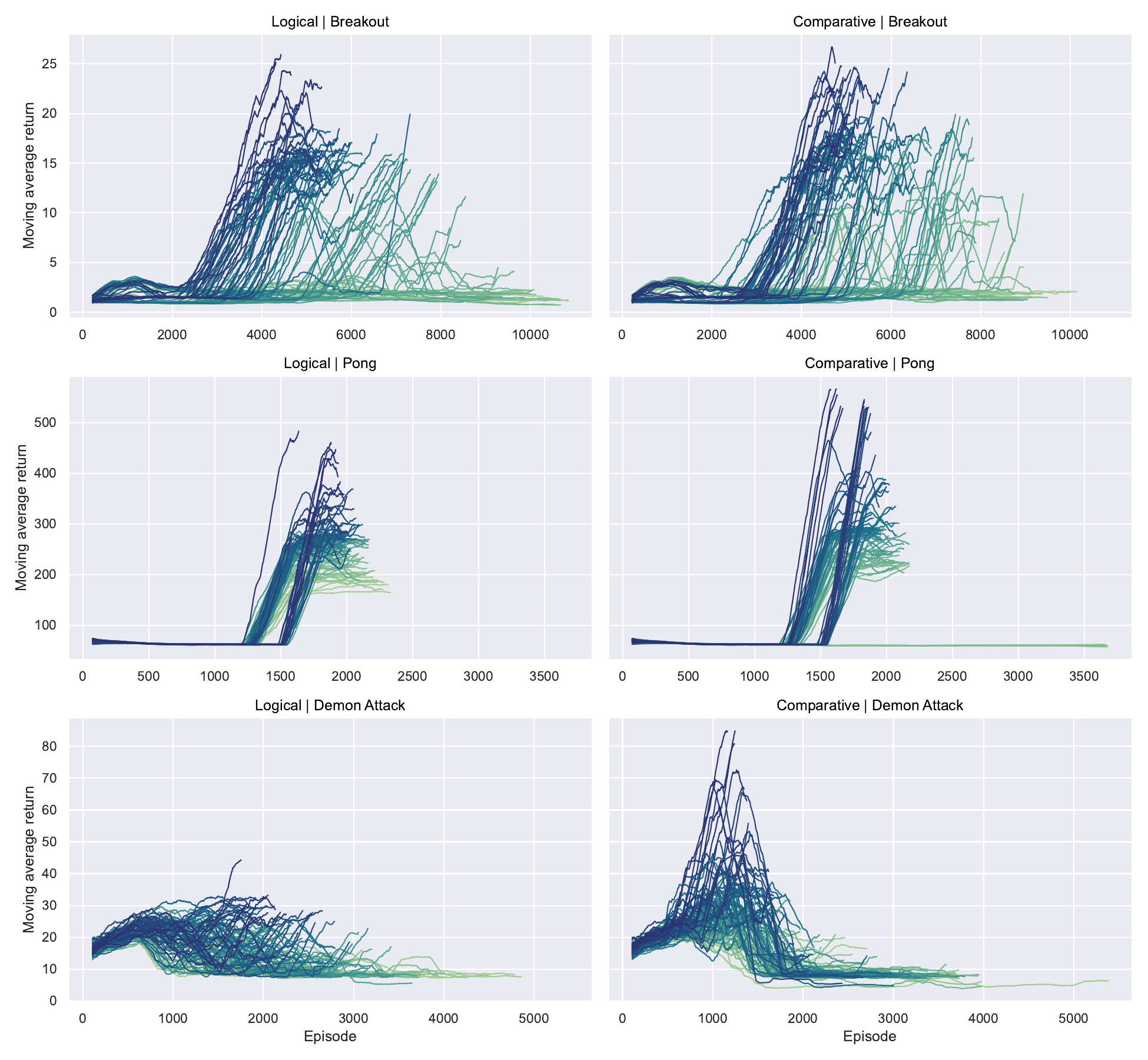}
    \caption{Training results by model version and game. The y-axis corresponds to the moving average of the human-normalized returns. For each game, the size of the moving window was set to the maximum number of training episodes divided by 12, and the minimum number of observations was set to the same number divided by 50. Each line corresponds to an individual random seed. Each line is colored by the returns' variance, with darker colors indicating higher variance.}
    \label{fig:train-results}
\end{figure*}

\begin{figure*}[ht!]
\centering
\includegraphics[width=0.7\linewidth]{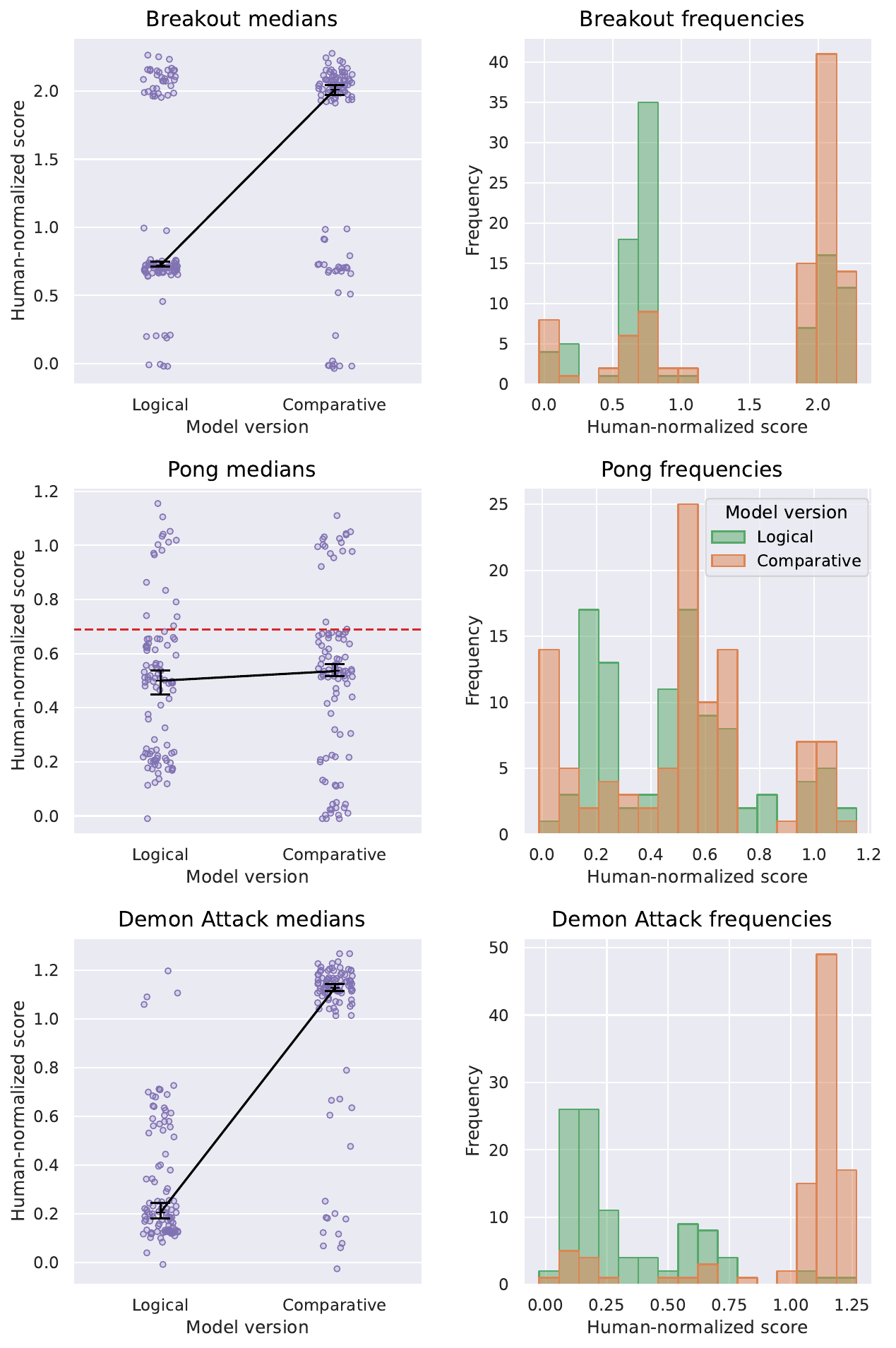}
\caption{Test results by model version and game. The first row shows the median performance as a line plot, along with the individual random-seed data points. Error bars are 95\% confidence intervals. The second row shows the corresponding frequency distributions.}\label{fig:results}
\end{figure*}

\begin{figure*}[ht!]
\centering
\includegraphics[width=\linewidth]{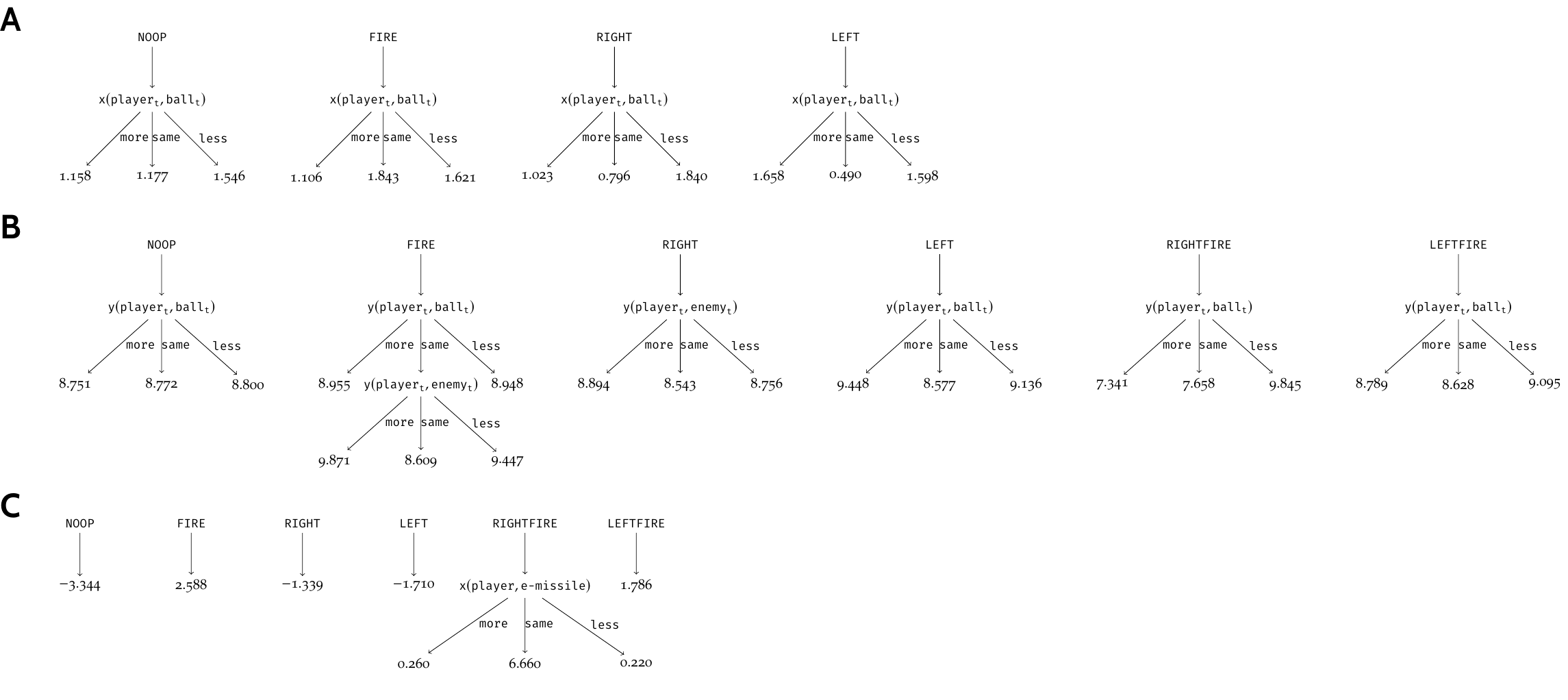}
\caption{Best state-action trees of RRTL (comparative version) on Breakout (Panel A), Pong (Panel B), and Demon Attack (Panel C). See text for details.}\label{fig:besttrees}
\end{figure*}

\subsection{Training}

For each game, we trained 100 random seeds of the logical and comparative versions of RRTL. Regarding the RL settings, $\alpha$ was set to 0.025 for Breakout and Pong and to 0.1 in Demmon Attack. For all games, we transformed the reward using the sign function, set $\gamma$ to 0.99, and $\epsilon$ was decayed from 1.0 to 0.1 multiplicatively over 500,000 steps. We trained the 100 random seeds for 2 million iterations in Breakout and Pong and for 3 million iterations in Demon Attack. Regarding the RRTL free parameters, for all games we set the maximum tree depth to 10, the minimal sample size to 100,000 and the significance level to 0.0001. In Breakout the RL agents used an action buffer that stored the last 10 actions and checked whether the agent had taken the same action 10 times consecutively, in which case the current action was uniformly sampled from the action space. During training, we checked at the end of each episode whether the episode return was bigger than the previous best return, in which case we saved the agent's set of state-action trees. The last saved set became the agent test trees.

In Pong, the enemy follows the ball by default ($r_{y(\texttt{ball}), y(\texttt{enemy})}=0.84$ in 10 random games). This has the effect of making the reward signal very sparse for an agent that follows a random policy (as is any RL agent following a $\epsilon$-greedy policy at the beginning of learning). To address this issue, we added $0.1$ to the reward at each time step to encourage the agent to play for as long as possible.

\section{Results}

\subsection{Training}

\autoref{fig:train-results} presents the learning trajectories of the 100 random seeds of the logical and comparative versions of RRTL for each game. In this plot, the x-axis corresponds to the number of training episodes, and the y-axis corresponds to the moving average of the training returns. For each game, the size of the moving average window was set to the maximum number of training episodes divided by 12, and the minimum number of observations for the window was set to the same number divided by 50. Each line corresponds to the learning trajectory of an individual random seed and is colored by the returns' variance, with darker colors indicating higher variance. 

The training results on Breakout are presented in the first row of \autoref{fig:train-results}. As shown, more seeds showed an uninterrupted rise in returns during learning in the logical version of the model than in the comparative version. This was due to the comparative version having more seeds whose returns destabilized after an initial period of successful learning. This trend exemplifies one of RRTL's current limitations: it cannot undo suboptimal splits made during learning. Another notable trend is that more seeds showed a flat learning curve in the logical version than in the comparative version. 

The second row of \autoref{fig:train-results} shows the training results on Pong. As shown, more seeds with higher overall returns were observed during learning in the comparative version of RRTL. In both versions of the model, several suboptimal splits compromised the learning trajectories. In contrast with Breakout, more seeds exhibited a flat learning curve in the comparative version of the model compared to the logical version. 

The training results on Demon Attack are presented in the third row of \autoref{fig:train-results}. As shown, there was a clear difference in the learning trajectories of the seeds between the two model versions. In the comparative version of RRTL, more seeds achieved higher returns overall; at the same time, however, performance tended to degrade after the initial phase of successful learning. As in Breakout, more seeds had flatter learning curves in the logical version of RRTL.

\subsection{Test}

To evaluate the test performance of the logical and comparative versions of RRTL, we tested the greedy policy\footnote{Note that, as aforementioned, even when following a greedy policy, our agents included a stochastic component in response to \say{empty} states.} of each of the 100 random seeds of both versions of the model on 100 test games. The performance of each seed was the human-normalized average return on these 100 test games. For the normalization, we used the human scores reported by \citet{mnih2015human}. \autoref{fig:results} presents the overall performance of RRTL by model version and game, along with the corresponding frequency distributions.

The results on Breakout are presented in the first row of \autoref{fig:results}. As shown in the first column, the majority of seeds in both model versions achieved performance above the random level. Furthermore, the maximum average return was similar in both model versions. However, the comparative version of RRTL showed a higher median performance ($2.011$) than the logical version ($0.722$). A Mann–Whitney U test found a statistically significant difference between the logical and comparative versions ($U=6374.5$, $p<0.05$, $\text{RBC}=0.275$). This advantage occurred because the results were much more consistent in the comparative version. For example, $70$ seeds achieved human-level or higher performance in the comparative version compared to only $35$ seeds in the logical version. These differences in the distribution of average returns are patent in the frequency plot in the second column. \autoref{fig:besttrees}A shows the state-action trees of the best random seed of the comparative version of RRTL. As can be observed by inspecting the state-action values, in this seed, the RRTL agent learned to follow the ball: it goes left if the player is to the right of the ball, it goes right if the player is to the left of the ball, and does not move if the player and the ball are at the same x-position.

The results on Pong are presented in the second row of \autoref{fig:results}. As shown in the first column, there were more random seeds with performance around random behavior level for both versions of the model compared to Breakout. Nonetheless, the majority of the seeds achieved performance above the random behavior level. Again, the maximum average return was similar in both model versions. Furthermore, the comparative version of RRTL showed an equivalent median performance ($0.535$) to the logical version ($0.5$). A Mann–Whitney U test did not find a statistically significant difference between the logical and comparative versions ($U=5339.0$, $p>0.05$, $\text{RBC}=0.068$). \autoref{fig:besttrees}B shows the state-action trees of the best random seed of the comparative version of RRTL. As can be seen by inspecting the state-action values, in this seed, the RRTL agent also learned to follow the ball: it goes up if the player is below the ball, it goes down if the player is above the ball, and does not move if the player and the ball are at the same y-position. Importantly, in this game, RRTL was able to select the critical $\texttt{y}(\texttt{player}_{\texttt{t}}, \texttt{ball}_{\texttt{t}})$ relation from a larger set of candidate relations in comparison to Breakout.

The results on Demon Attack are presented in the third row of \autoref{fig:results}. As shown in the first column, while the majority of seeds in the comparative version achieved performance well above the random behavior level, the majority of seeds in the logical version were close to the random behavior level. Again, the maximum average return was similar in both model versions. However, the comparative version of RRTL showed a higher median performance ($1.127$) than the logical version ($0.206$). A Mann–Whitney U test found a statistically significant difference between the logical and comparative versions ($U=8962.0$, $p<0.05$, $\text{RBC}=0.792$). This advantage occurred because the results were much more consistent in the comparative version. For example, $83$ seeds achieved human-level or higher performance in the comparative version compared to only $4$ seeds in the logical version. These differences in the distribution of average returns are patent in the frequency plot in the second column. \autoref{fig:besttrees}C shows the state-action trees of the best random seed of the comparative version of RRTL. As can be seen, in this case the RRTL agent selected the $\texttt{x}(\texttt{player}, \texttt{e-missile})$ relation, even though it made a split only for the \texttt{RIGHTFIRE} action. In this seed, RRTL learned a simple policy that roughly amounts to avoiding the enemy missile while shooting as much as possible. In particular, in this seed, the RRTL agent stays put and fires if the player is to the left of the enemy missile, goes right and fires if the player and the enemy missile are at the same x-position, and stays put and fires if the player is to the right of the enemy missile.

\section{Discussion}

In this work, we studied how to select appropriate relational representations to build a policy when a large vocabulary of relations is available to describe the environment's state. Using a function approximator developed in RRL, we developed RRTL, a model that learns ground relational policies by making ternary splits based on the comparative values \emph{more}, \emph{same}, and \emph{less} that characterize comparative relations like \say{above} or \say{bigger-than}. We tested our model in three Atari games\textemdash Breakout, Pong, and Demon Attack\textemdash that involved an increasing number of potential relations. In each case, RRTL built simple relational policies based on a selected set of a few relevant relations. RRTL can be considered a proof-of-principle of the idea of using reinforcement to learn which of all available relations are relevant to characterize the task at hand.

We hypothesized that, while in principle logical and comparative splits can express the same policies, comparative splits should lead to more robust learning because of (1) the Q-values of the \say{more}, \say{less} and \say{same} partitions induced by comparative relations are likely related for most environments and (2) the higher number of splits in the logical case entail more chances of error during learning. Our results indeed showed that making comparative splits improved RRTL's performance. Specifically, in two out of the three games tested, the comparative version of the model had fewer random seeds showing suboptimal policies. Furthermore, in two of the three games tested, there were more random seeds at the maximum level of performance in the comparative version of the model. 

While in this work we showed that RRTL can learn simple relational policies incrementally, building more human-like relational policies requires using explicit representations of physics, agents, events, and goals \citep{tsividis2021human,pouncy2022inductive}. In our simulations, this point was clear in Pong, where representing the position of the player relative to the ball was not sufficient to consistently beat the opponent. Consistently winning in Pong likely requires representing the ball's trajectory with greater granularity, including its acceleration, to predict its future position. We think that building complex relational world models incrementally, without fully pre-specifying these theories beforehand, is a central problem for cognitive science \citep{doumas2022theory}. By introducing a reinforcement-based method for selecting relevant relational representations tailored to specific tasks, we think that RRTL offers a foundational step toward addressing this challenge.

In contrast to the variabilized rules learned by RRL systems developed in AI \citep[e.g.,][]{driessens2001speeding,guestrin2003generalizing,pasula2007learning}, RRTL learns policies composed of ground rules. We propose that, instead of learning variabilized rules directly through interaction with the environment, humans \say{lift} initially ground policies to a representation more akin to a first-order relational policy via schema induction. This process involves extracting the shared relational structure through analogical comparison and mapping across two different situations \citep{hummel2003symbolic,gick1983schema}. We think this process could be used to compare different instances of a ground policy (e.g., several random seeds of RRTL) to isolate its relational content and abstract away the specific objects attached to it. Furthermore, schema induction could be used to compare relational policies of different tasks (e.g., Breakout and Pong) to form a more abstract representation of their common attributes. Relatedly, Foster and Jones \citet{foster2017reinforcement} have shown that it is possible to use RL to \emph{guide} the induction of useful schemas to solve a task. More generally, we think that integrating RRL, schema induction, and analogical inference has the potential to advance our understanding of how we build relational theories of the environment and use this knowledge to generalize flexibly and rapidly across domains.

In its current form, RRTL has some limitations. The first one concerns the frequentist approach used to select the relation candidates for splitting state-action trees. Because frequentist statistical tests are highly sensitive to the sample size, all candidates should have a similar number of visits to be comparable. This is a problem in any realistic environment, since there can be relations that apply only rarely but nonetheless have a large impact on the optimal policy. For example, imagine a version of Demon Attack where the small enemies descend onto the spaceship too fast to try to shoot them from bellow, in which case the best option would be to avoid the small enemies and, consequently, the $\texttt{x}(\texttt{player}, \texttt{e-small-1})$ relation would be very important, but only so when there is a small enemy on the screen. We think that, in this case, a Bayesian approach to candidate selection would be useful for accounting for this kind of phenomenon. A related limitation is that, as seen in the training results, RRTL currently lacks a mechanism to reverse suboptimal splits made during learning, which can significantly compromise its learning trajectory. Fortunately, \cite{ramon2007transfer} developed a set of tree-restructuring operations for partial policy transfer on RRL. We think that combining these operations with a Bayesian approach to splitting candidate selection is a promising future direction for RRTL.

Another direction for future research is to develop a more refined representation of actions. In this work, we treated actions as atomic and independent primitives. However, the actions in the games we tested were clearly related. For example, in Breakout, if a split is relevant for the action \texttt{LEFT}, it is probably also important for the action \texttt{RIGHT}. This relationship could be captured by representing actions as their consequences in the environment. In this approach, the action \texttt{LEFT} would become \texttt{less-x}(\texttt{player}$_{\texttt{t}}$, \texttt{player}$_{\texttt{t-1}}$), \texttt{NOOP} would become \texttt{same-x}(\texttt{player}$_{\texttt{t}}$, \texttt{player}$_{\texttt{t-1}}$), and \texttt{RIGHT} would become \texttt{more-x}(\texttt{player}$_{\texttt{t}}$, \texttt{player}$_{\texttt{t-1}}$). Representing actions in this way could allow RRTL to split all directly related actions simultaneously, potentially improving its learning robustness.

Since our motivation was to build a RRL model that can interface with analogical inference and previous computational modeling work has shown that structured relational representations are necessary to perform analogy \cite[e.g.,][]{doumas2008theory,doumas2022theory,hummel2003symbolic}, in this work we used the relational regression tree developed by \cite{driessens2001speeding} to approximate the state-action value function. However, it is worth noting that recently several deep neural network models of RRL have been proposed \cite[e.g.,][]{jiang2019neural,dong2018neural,zimmer2021differentiable,beretta2023preliminary}. In general, these models define a \emph{soft logic}\textemdash i.e., one with truth values falling into the $[0,1]$ interval\textemdash and a series of differentiable gating operations that allow to approximate the behaviour of a logic program. While these models are an impressive demonstration of the possibility of combining symbolic computation and deep learning in AI, it is not clear at the moment whether it is possible to integrate these function approximators with models of relational reasoning and analogical inference. We think that building an efficient and neurally plausible function approximator suited to work with structured relational representations is an important goal for RRL research.

In conclusion, RRL provides a computational framework for understanding how the cognitive system identifies task-relevant relational representations and for building relational policies incrementally. We hope that the present work helps to stimulate further cognitive science research in this area.

\bibliographystyle{apacite}
\bibliography{cas-refs.bib}

\onecolumn
\appendix
\section{State representations for Pong and Demon Attack}
\label{app1}
\begin{table*}[h]
    \centering
    \caption{Pong State Representation.}
    \label{tab:PongTable}
    \begin{tabular}{@{}cllll@{}}         \toprule
    & & & \multicolumn{2}{c}{Relation} \\
    \cmidrule(l){4-5}
    Dim & Obj-1 & Obj-2 & Logical & Comparative \\ \midrule
    
    \texttt{x} & $\texttt{player}_{\texttt{t}}$ & $\texttt{ball}_{\texttt{t}}$ & $\texttt{more-x}(\texttt{player}_{\texttt{t}}$, $\texttt{ball}_{\texttt{t}})$ & $\texttt{x}(\texttt{player}_{\texttt{t}}, \texttt{ball}_{\texttt{t}})$\\
    
    & & & $\texttt{same-x}(\texttt{player}_{\texttt{t}}$, $\texttt{ball}_{\texttt{t}})$ & \\
    
    & & & $\texttt{less-x}(\texttt{player}_{\texttt{t}}, \texttt{ball}_{\texttt{t}})$ & \\ \midrule
    
    \texttt{x} & $\texttt{player}_{\texttt{t}}$ & $\texttt{enemy}_{\texttt{t}}$ & $\texttt{more-x}(\texttt{player}_{\texttt{t}}, \texttt{enemy}_{\texttt{t}})$ & $\texttt{x}(\texttt{player}_{\texttt{t}}, \texttt{enemy}_{\texttt{t}})$\\
    
    & & & $\texttt{same-x}(\texttt{player}_{\texttt{t}}, \texttt{enemy}_{\texttt{t}})$ & \\
    
    & & & $\texttt{less-x}(\texttt{player}_{\texttt{t}}, \texttt{enemy}_{\texttt{t}})$ & \\ \midrule
    
    \texttt{x} & $\texttt{ball}_{\texttt{t}}$ & $\texttt{enemy}_{\texttt{t}}$ & $\texttt{more-x}(\texttt{ball}_{\texttt{t}}, \texttt{enemy}_{\texttt{t}})$ & $\texttt{x}(\texttt{ball}_{\texttt{t}}, \texttt{enemy}_{\texttt{t}})$\\
    
    & & & $\texttt{same-x}(\texttt{ball}_{\texttt{t}}, \texttt{enemy}_{\texttt{t}})$ & \\
    
    & & & $\texttt{less-x}(\texttt{ball}_{\texttt{t}}, \texttt{enemy}_{\texttt{t}})$ & \\ \midrule

    \texttt{y} & $\texttt{player}_{\texttt{t}}$ & $\texttt{ball}_{\texttt{t}}$ & $\texttt{more-y}(\texttt{player}_{\texttt{t}}, \texttt{ball}_{\texttt{t}})$ & $\texttt{y}(\texttt{player}_{\texttt{t}}, \texttt{ball}_{\texttt{t}})$\\
    
    & & & $\texttt{same-y}(\texttt{player}_{\texttt{t}}, \texttt{ball}_{\texttt{t}})$ & \\
    
    & & & $\texttt{less-y}(\texttt{player}_{\texttt{t}}, \texttt{ball}_{\texttt{t}})$ & \\ \midrule
    
    \texttt{y} & $\texttt{player}_{\texttt{t}}$ & $\texttt{enemy}_{\texttt{t}}$ & $\texttt{more-y}(\texttt{player}_{\texttt{t}}, \texttt{enemy}_{\texttt{t}})$ & $\texttt{y}(\texttt{player}_{\texttt{t}}, \texttt{enemy}_{\texttt{t}})$\\
    
    & & & $\texttt{same-y}(\texttt{player}_{\texttt{t}}, \texttt{enemy}_{\texttt{t}})$ & \\
    
    & & & $\texttt{less-y}(\texttt{player}_{\texttt{t}}, \texttt{enemy}_{\texttt{t}})$ & \\ \midrule
    
    \texttt{y} & $\texttt{ball}_{\texttt{t}}$ & $\texttt{enemy}_{\texttt{t}}$ & $\texttt{more-y}(\texttt{ball}_{\texttt{t}}, \texttt{enemy}_{\texttt{t}})$ & $\texttt{x}(\texttt{ball}_{\texttt{t}}, \texttt{enemy}_{\texttt{t}})$\\
    
    & & & $\texttt{same-y}(\texttt{ball}_{\texttt{t}}, \texttt{enemy}_{\texttt{t}})$ & \\
    
    & & & $\texttt{less-y}(\texttt{ball}_{\texttt{t}}, \texttt{enemy}_{\texttt{t}})$ & \\ \midrule
    
    \texttt{x} & $\texttt{ball}_{\texttt{t}}$ & $\texttt{ball}_{\texttt{t-1}}$ & $\texttt{more-x}(\texttt{ball}_{\texttt{t}}, \texttt{ball}_{\texttt{t-1}})$ & $\texttt{x}(\texttt{ball}_{\texttt{t}}, \texttt{ball}_{\texttt{t-1}})$ \\
    
    & & & $\texttt{same-x}(\texttt{ball}_{\texttt{t}}, \texttt{ball}_{\texttt{t-1}})$ & \\
    
    & & & $\texttt{less-x}(\texttt{ball}_{\texttt{t}}, \texttt{ball}_{\texttt{t-1}})$ & \\ \midrule
    
    \texttt{y} & $\texttt{ball}_{\texttt{t}}$ & $\texttt{ball}_{\texttt{t-1}}$ & $\texttt{more-y}(\texttt{ball}_{\texttt{t}}, \texttt{ball}_{\texttt{t-1}})$ & $\texttt{y}(\texttt{ball}_{\texttt{t}}, \texttt{ball}_{\texttt{t-1}})$ \\
    
    & & & $\texttt{same-y}(\texttt{ball}_{\texttt{t}}, \texttt{ball}_{\texttt{t-1}})$ & \\
    
    & & & $\texttt{less-y}(\texttt{ball}_{\texttt{t}}, \texttt{ball}_{\texttt{t-1}})$ & \\ \midrule
    
    \texttt{y} & $\texttt{enemy}_{\texttt{t}}$ & $\texttt{enemy}_{\texttt{t-1}}$ & $\texttt{more-y}(\texttt{enemy}_{\texttt{t}}, \texttt{enemy}_{\texttt{t-1}})$ & $\texttt{y}(\texttt{enemy}_{\texttt{t}}, \texttt{enemy}_{\texttt{t-1}})$ \\
    
    & & & $\texttt{same-y}(\texttt{enemy}_{\texttt{t}}, \texttt{enemy}_{\texttt{t-1}})$ & \\
    
    & & & $\texttt{less-y}(\texttt{enemy}_{\texttt{t}}, \texttt{enemy}_{\texttt{t-1}})$ & \\ \midrule
    
    & $\texttt{player}_{\texttt{t}}$ & $\texttt{ball}_{\texttt{t}}$ & $\texttt{in-contact}(\texttt{player}_{\texttt{t}}, \texttt{ball}_{\texttt{t}})$ & \\ \midrule
    
    & $\texttt{ball}_{\texttt{t}}$ & $\texttt{enemy}_{\texttt{t}}$ & $\texttt{in-contact}(\texttt{ball}_{\texttt{t}}, \texttt{enemy}_{\texttt{t}})$ & \\
    \bottomrule
    \end{tabular}
\end{table*}

\newpage

{

\begin{longtable}{@{}cllll@{}}    
    \caption{Demon Attack State Representation}
    \label{tab:DemonAttackTable}\\\toprule
    & & & \multicolumn{2}{c}{Relation} \\
    \cmidrule(l){4-5}
    Dim & Obj-1 & Obj-2 & Logical & Comparative \\ \midrule
    \endfirsthead
    \caption* {\textnormal{\textbf{Table A2 (continued)}}\\
    \vspace{10pt}
    Demon Attack State Representation.}\\\toprule
    & & & \multicolumn{2}{c}{Relation} \\
    \cmidrule(l){4-5}
    Dim & Obj-1 & Obj-2 & Logical & Comparative \\ \midrule
    \endhead
    \endfoot
    \bottomrule
    \endlastfoot
    
    \texttt{x} & $\texttt{player}$ & $\texttt{e-missile}$ & $\texttt{more-x}(\texttt{player}, \texttt{e-missile})$ & $\texttt{x}(\texttt{player}, \texttt{e-missile})$\\
    
    & & & $\texttt{same-x}(\texttt{player}, \texttt{e-missile})$ & \\
    
    & & & $\texttt{less-x}(\texttt{player}, \texttt{e-missile})$ & \\ \midrule
    
    \texttt{x} & $\texttt{player}$ & $\texttt{e-big-1}$ & $\texttt{more-x}(\texttt{player}, \texttt{e-big-1})$ & $\texttt{x}(\texttt{player}, \texttt{e-big-1})$\\
    
    & & & $\texttt{same-x}(\texttt{player}, \texttt{e-big-1})$ & \\
    
    & & & $\texttt{less-x}(\texttt{player}, \texttt{e-big-1})$ & \\ \midrule
    
    \texttt{x} & $\texttt{player}$ & $\texttt{e-big-2}$ & $\texttt{more-x}(\texttt{player}, \texttt{e-big-2})$ & $\texttt{x}(\texttt{player}, \texttt{e-big-2})$\\
    
    & & & $\texttt{same-x}(\texttt{player}, \texttt{e-big-2})$ & \\
    
    & & & $\texttt{less-x}(\texttt{player}, \texttt{e-big-2})$ & \\ \midrule
    
    \texttt{x} & $\texttt{player}$ & $\texttt{e-big-3}$ & $\texttt{more-x}(\texttt{player}, \texttt{e-big-3})$ & $\texttt{x}(\texttt{player}, \texttt{e-big-3})$\\
    
    & & & $\texttt{same-x}(\texttt{player}, \texttt{e-big-3})$ & \\
    
    & & & $\texttt{less-x}(\texttt{player}, \texttt{e-big-3})$ & \\ \midrule

    \texttt{x} & $\texttt{player}$ & $\texttt{e-small-1}$ & $\texttt{more-x}(\texttt{player}, \texttt{e-small-1})$ & $\texttt{x}(\texttt{player}, \texttt{e-small-1})$\\
    
    & & & $\texttt{same-x}(\texttt{player}, \texttt{e-small-1})$ & \\
    
    & & & $\texttt{less-x}(\texttt{player}, \texttt{e-small-1})$ & \\ \midrule
    
    \texttt{x} & $\texttt{player}$ & $\texttt{e-small-2}$ & $\texttt{more-x}(\texttt{player}, \texttt{e-small-2})$ & $\texttt{x}(\texttt{player}, \texttt{e-small-2})$\\
    
    & & & $\texttt{same-x}(\texttt{player}, \texttt{e-small-2})$ & \\
    
    & & & $\texttt{less-x}(\texttt{player}, \texttt{e-small-2})$ & \\ \midrule

    \texttt{x} & $\texttt{player}$ & $\texttt{e-small-3}$ & $\texttt{more-x}(\texttt{player}, \texttt{e-small-3})$ & $\texttt{x}(\texttt{player}, \texttt{e-small-3})$\\
    
    & & & $\texttt{same-x}(\texttt{player}, \texttt{e-small-3})$ & \\
    
    & & & $\texttt{less-x}(\texttt{player}, \texttt{e-small-3})$ & \\ \midrule
    
    \texttt{x} & $\texttt{player}$ & $\texttt{e-small-4}$ & $\texttt{more-x}(\texttt{player}, \texttt{e-small-4})$ & $\texttt{x}(\texttt{player}, \texttt{e-small-4})$\\
    
    & & & $\texttt{same-x}(\texttt{player}, \texttt{e-small-4})$ & \\
    
    & & & $\texttt{less-x}(\texttt{player}, \texttt{e-small-5})$ & \\ \midrule
    
    \texttt{x} & $\texttt{player}$ & $\texttt{e-small-5}$ & $\texttt{more-x}(\texttt{player}, \texttt{e-small-5})$ & $\texttt{x}(\texttt{player}, \texttt{e-small-5})$\\
    
    & & & $\texttt{same-x}(\texttt{player}, \texttt{e-small-5})$ & \\
    
    & & & $\texttt{less-x}(\texttt{player}, \texttt{e-small-5})$ & \\ \midrule

    \texttt{x} & $\texttt{player}$ & $\texttt{e-small-6}$ & $\texttt{more-x}(\texttt{player}, \texttt{e-small-6})$ & $\texttt{x}(\texttt{player}, \texttt{e-small-6})$\\
    
    & & & $\texttt{same-x}(\texttt{player}, \texttt{e-small-6})$ & \\
    
    & & & $\texttt{less-x}(\texttt{player}, \texttt{e-small-6})$ & \\ \midrule

    \texttt{y} & $\texttt{player}$ & $\texttt{e-missile}$ & $\texttt{more-y}(\texttt{player}, \texttt{e-missile})$ & $\texttt{y}(\texttt{player}, \texttt{e-missile})$\\
    
    & & & $\texttt{same-y}(\texttt{player}, \texttt{e-missile})$ & \\
    
    & & & $\texttt{less-y}(\texttt{player}, \texttt{e-missile})$ & \\ \midrule

    \texttt{y} & $\texttt{player}$ & $\texttt{e-big-1}$ & $\texttt{more-y}(\texttt{player}, \texttt{e-big-1})$ & $\texttt{y}(\texttt{player}, \texttt{e-big-1})$\\
    
    & & & $\texttt{same-y}(\texttt{player}, \texttt{e-big-1})$ & \\
    
    & & & $\texttt{less-y}(\texttt{player}, \texttt{e-big-1})$ & \\ \midrule

    \texttt{y} & $\texttt{player}$ & $\texttt{e-big-2}$ & $\texttt{more-y}(\texttt{player}, \texttt{e-big-2})$ & $\texttt{y}(\texttt{player}, \texttt{e-big-2})$\\
    
    & & & $\texttt{same-y}(\texttt{player}, \texttt{e-big-2})$ & \\
    
    & & & $\texttt{less-y}(\texttt{player}, \texttt{e-big-2})$ & \\ \midrule
    
    \texttt{y} & $\texttt{player}$ & $\texttt{e-big-3}$ & $\texttt{more-y}(\texttt{player}, \texttt{e-big-3})$ & $\texttt{y}(\texttt{player}, \texttt{e-big-3})$\\
    
    & & & $\texttt{same-y}(\texttt{player}, \texttt{e-big-3})$ & \\
    
    & & & $\texttt{less-y}(\texttt{player}, \texttt{e-big-3})$ & \\ \midrule
    \pagebreak

    \texttt{y} & $\texttt{player}$ & $\texttt{e-small-1}$ & $\texttt{more-y}(\texttt{player}, \texttt{e-small-1})$ & $\texttt{y}(\texttt{player}, \texttt{e-small-1})$\\
    
    & & & $\texttt{same-y}(\texttt{player}, \texttt{e-small-1})$ & \\
    
    & & & $\texttt{less-y}(\texttt{player}, \texttt{e-small-1})$ & \\ \midrule

    \texttt{y} & $\texttt{player}$ & $\texttt{e-small-2}$ & $\texttt{more-y}(\texttt{player}, \texttt{e-small-2})$ & $\texttt{y}(\texttt{player}, \texttt{e-small-2})$\\
    
    & & & $\texttt{same-y}(\texttt{player}, \texttt{e-small-2})$ & \\
    
    & & & $\texttt{less-y}(\texttt{player}, \texttt{e-small-2})$ & \\ \midrule
    
    \texttt{y} & $\texttt{player}$ & $\texttt{e-small-3}$ & $\texttt{more-y}(\texttt{player}, \texttt{e-small-3})$ & $\texttt{y}(\texttt{player}, \texttt{e-small-3})$\\
    
    & & & $\texttt{same-y}(\texttt{player}, \texttt{e-small-3})$ & \\
    
    & & & $\texttt{less-y}(\texttt{player}, \texttt{e-small-3})$ & \\ \midrule
    
    \texttt{y} & $\texttt{player}$ & $\texttt{e-small-4}$ & $\texttt{more-y}(\texttt{player}, \texttt{e-small-4})$ & $\texttt{y}(\texttt{player}, \texttt{e-small-4})$\\
    
    & & & $\texttt{same-y}(\texttt{player}, \texttt{e-small-4})$ & \\
    
    & & & $\texttt{less-y}(\texttt{player}, \texttt{e-small-5})$ & \\ \midrule

    \texttt{y} & $\texttt{player}$ & $\texttt{e-small-5}$ & $\texttt{more-y}(\texttt{player}, \texttt{e-small-5})$ & $\texttt{y}(\texttt{player}, \texttt{e-small-5})$\\
    
    & & & $\texttt{same-y}(\texttt{player}, \texttt{e-small-5})$ & \\
    
    & & & $\texttt{less-y}(\texttt{player}, \texttt{e-small-5})$ & \\ \midrule

    \texttt{y} & $\texttt{player}$ & $\texttt{e-small-6}$ & $\texttt{more-y}(\texttt{player}, \texttt{e-small-6})$ & $\texttt{y}(\texttt{player}, \texttt{e-small-6})$\\
    
    & & & $\texttt{same-y}(\texttt{player}, \texttt{e-small-6})$ & \\
    
    & & & $\texttt{less-y}(\texttt{player}, \texttt{e-small-6})$ & \\

\end{longtable}
} 

\end{document}